\documentclass[]{ceurart}
\usepackage{listings}
\begin{document}

\copyrightyear{2026}
\copyrightclause{Copyright for this paper by its authors.
  Use permitted under Creative Commons License Attribution 4.0
  International (CC BY 4.0).}

\conference{CLEF 2026 Working Notes, 21 -- 24 September 2026, Jena, Germany}

\title{DS@GT ARC at MEDIQA-CORE-Task-1 2026: Trimodal Model Fusion with Task-Specific Gates for Brain Tumor Subtype Classification}

\author[1]{Hoang Thanh Thanh Truong}[
    orcid=0009-0007-4130-3349,
    email=htruong47@gatech.edu
]
\cormark[1]

\author[1,2]{Charles R. Clark}[
    orcid=0009-0008-0415-111X,
    email=cclark339@gatech.edu
]

\address[1]{Georgia Institute of Technology, North Ave NW, Atlanta, GA 30332}
\address[2]{University of Florida, Stadium Rd, Gainesville, FL 32611}
\cortext[1]{Corresponding author.}

\begin{abstract}
Brain tumor diagnosis is a time-sensitive process in which patients may wait 
weeks for a finalized pathology report. This problem motivates automated 
systems that classify tumor subtype from multimodal inputs. This paper details the DS@GT ARC team's work for ImageCLEFmed MEDIQA-CORE 2026 Task~1, Brain Tumor Subtype Classification. The task evaluates three glioma classification problems: Level-1 Molecular Type, LGG vs HGG, and WHO Grade. We combine pre-extracted MRI (NeuroVFM) and histopathology (Prov-GigaPath) embeddings with free-text radiology reports. Our team explored two trimodal fusion architectures, two report encoders (RadBERT and Llama-3.1-8B-Instruct), and a biologically motivated post-processing stage. We achieve a mean macro-F1 of 0.801 under the Fully Multimodal condition, exceeding the organizers' baseline of 0.796 and ranking second among the teams whose code passed verification. Additional evaluation across modality-dropping conditions shows that this advantage depends heavily on the availability of the histopathology modality, and that our system falls behind the baseline when modalities are missing. Our 
code is available on GitHub at  https://github.com/dsgt-arc/imageclef-mediqacore-2026.
\end{abstract}


\maketitle

\section{Introduction}

Time matters for patients with brain tumors. Every week between symptom onset and definitive diagnosis is a week before treatment can begin. Despite modern imaging methods, patients with brain tumors could wait a median of 39 days from first symptom to diagnosis~\cite{alther2020firstsymptoms}. For the most aggressive primary brain tumor in adults, glioblastoma, this delay is particularly costly. Their tumors could grow at a median rate of 1.4\% per day, with nearly one-third doubling in volume between diagnostic and preoperative scans~\cite{stensjoen2015growth}.

Reducing this delay is therefore a critical goal. To address it, 
we must first understand the multi-step, multimodal nature of the diagnostic 
workflow. A patient typically begins with magnetic resonance imaging (MRI) scan, proceeds to histopathologic examination, and finally awaits a diagnostic report from a board-certified neuropathologist ~\cite{CoRe-BT-arXiv}. Each step takes 
time and requires a different specialist. This calls for automated multimodal 
classification systems to shorten the pipeline. By producing preliminary tumor subtype predictions 
from whichever modalities are available at a given point in the workflow, 
such systems can inform downstream clinical decisions during the wait for the final expert diagnosis.

The MEDIQA-CORE 2026 Brain Tumor Subtype Classification task~\cite{mediqa-core-task-1, ImageCLEF2026} evaluates this idea on the CoRe-BT dataset~\cite{CoRe-BT-arXiv}. The dataset provides paired MRI embeddings, whole-slide histopathology embeddings, and 
free-text radiology reports for glioma patients. Participants are evaluated on 
three classification tasks: Level-1 Molecular Type (a three-class molecular 
grouping), LGG vs HGG (a binary low-grade versus high-grade distinction), and 
WHO Grade (a three-class grading task aligned with the 2021 WHO CNS 
classification~\cite{louis2021who}). Although the three tasks share the same inputs, they capture different aspects of tumor biology, so each may benefit from a different combination of the modalities.

In this work, we explore two trimodal fusion strategies for jointly predicting 
the three tasks. The first is a cross-modal self-attentive fusion. The 
three modality latents attend to one another through a shared transformer trunk 
before a single learned gate produces the fused representation. 

The second, which is our primary submitted system, replaces the single shared gate with task-specific gates. Rather than fusing the three modality embeddings 
into one shared representation, the model maintains three independent learned 
gates, one per classification task, so that each task can learn its own optimal 
modality mixing. 

The MRI and histopathology embeddings are provided pre-computed by the challenge organizers, extracted with NeuroVFM~\cite{kondepudi2025neurovfm} 
and Prov-GigaPath~\cite{xu2024gigapath} respectively. We extract dense report 
embeddings using Llama-3.1-8B-Instruct~\cite{grattafiori2024llama}. We train via 5-fold stratified cross-validation, ensemble fold predictions via logit 
averaging. Then, we apply biologically motivated post-processing rules verified on the training set to enforce known constraints between task outputs. Our submitted system achieved a mean macro-F1 of 0.801 under the Fully Multimodal condition, exceeding the organizers' baseline of 0.796 and ranking second among the teams whose code passed verification~\cite{mediqa-core-task-1}. This advantage depends heavily on the availability of the histopathology modality and does not hold when modalities are missing.

\section{Related Work}

\subsection{Deep Learning for Glioma Subtype Classification}

Earlier work on glioma classification has fallen into two camps: MRI-only molecular prediction (e.g., IDH status from residual CNNs~\cite{Chang2018ResNetIDH}) and histology-based intraoperative diagnosis (e.g., real-time CNN classification on stimulated Raman histology~\cite{Hollon2020SRH} and DeepGlioma's molecular subtyping~\cite{Hollon2023DeepGlioma}). Both are single-modality and therefore cannot exploit the trimodal diagnostic workflow followed in practice. In parallel, oncological AI has shifted toward multimodal integration of heterogeneous data sources such as imaging, omics, EHRs, and clinical text, yielding richer representations of patient state than unimodal models~\cite{Acosta2022MultimodalBiomedicalAI} and improving the robustness and accuracy of diagnostic and prognostic predictions~\cite{Lipkova2022MultimodalAI}. PORPOISE~\cite{Chen2022PanCancerMultimodal}, for instance, fuses histology and genomics across 14 cancer types, including glioma, demonstrating consistent gains over single-modality baselines. The CoRe-BT benchmark~\cite{CoRe-BT-arXiv} brings this multimodal framing to brain-tumor typing specifically, releasing paired MRI, whole-slide histopathology, and free-text reports --- the setting we target in this work.

\subsection{Foundation Models as Frozen Modality Encoders}

\subsubsection{Pathology}

In recent years, a wave of foundation models has emerged for processing pathology data. Prov-GigaPath~\cite{xu2024gigapath} is a foundation model for whole-slide images (WSIs) that couples a ViT~\cite{Dosovitskiy2021ViT} tile encoder with a LongNet-based slide encoder to capture long-range, whole-slide context. It was trained on $\sim$1.3 billion tiles from $\sim$171 thousand WSIs ($\sim$30 thousand patients, 31 tissue types) sourced from a single US health network, and evaluated on 26 digital pathology tasks. This model was used to generate the WSI embeddings provided in the CoRe-BT dataset~\cite{CoRe-BT-arXiv}.

UNI~\cite{Chen2024PathFoundationModel} is a general-purpose, vision-only computational pathology (CPath) transformer~\cite{vaswani2017attention} foundation model trained on more than 100 million patches from over 100 thousand WSIs spanning 20 tissue types. Like Prov-GigaPath, it is an image-only backbone whose multimodal use is achieved through downstream fusion with other data.

CONCH~\cite{Lu2024VisionLanguagePathology} is, in contrast, a natively image--text pathology foundation model. Trained on $\sim$1.17 million image--caption pairs via CLIP-like~\cite{Radford2021CLIP} contrastive image--text pretraining, it achieves state-of-the-art performance on 14 vision--language tasks spanning classification, segmentation, retrieval, and captioning.

\subsubsection{Brain MRIs}

Neuroimaging has seen a parallel surge of foundation models. NeuroFM~\cite{Dibble2026NeuroFM} is a disease-naive, normative model trained exclusively on $\sim$100 thousand healthy \emph{synthetic} structural MRIs to predict morphometric and demographic targets; evaluated on $\sim$136 thousand real multi-cohort scans, it organizes structural MRIs into population-level patterns of brain health and supports applications such as early dementia-risk estimation. BrainIAC~\cite{Tak2026BrainIAC} is a general-purpose brain MRI model trained via self-supervised contrastive learning on nearly 49 thousand real, unlabeled MRIs to learn a transferable ``Brain Imaging Adaptive Core'' representation with strong few-shot performance. Both are image-only models that can be extended to multimodal settings through downstream fusion.

NeuroVFM~\cite{kondepudi2025neurovfm} is a clinical neuroimaging foundation model spanning both MRI and CT. Trained on $\sim$5.24 million uncurated MRI and CT volumes from a single health system using a volumetric joint-embedding predictive architecture (Vol-JEPA), it learns comprehensive representations of anatomy and pathology and achieves state-of-the-art performance on multiple diagnosis- and report-related tasks; it was used to generate the MRI embeddings provided in CoRe-BT~\cite{CoRe-BT-arXiv}.

Finally, RadFM~\cite{Wu2023RadFM} is a generalist radiological foundation model applicable to both 2- and 3-dimensional inputs. Trained on the custom MedMD ($\sim$16 million scans and text) and RadMD ($\sim$3 million radiology image--text pairs) datasets, it is a visually conditioned generative vision--language model that handles image and text natively and achieves state-of-the-art performance on RadBench relative to other multimodal foundation models.

\subsubsection{Clinical Text}

Since the early 2020s, large language models (LLMs) have become central to text-based tasks. General-purpose LLMs such as Llama~3~\cite{grattafiori2024llama} can be applied across a wide range of tasks, including as the frozen text feature extractor we use in this work.

Specialized medical LLMs have also begun to appear. Me-LLaMA~\cite{Xie2025MeLLaMA} is an open-source medical foundation LLM for broad text analysis, built from $\sim$129 billion pre-training tokens and $\sim$214 thousand instruction samples drawn from papers, guidelines, and EHR notes; it outperforms LLaMA and prior open medical LLMs in both zero-shot and supervised settings. Med-PaLM~2~\cite{Singhal2025ExpertMedicalQA} targets expert-level medical question answering (QA), developed through medical-domain fine-tuning of a proprietary base model (in contrast to the open Me-LLaMA) combined with prompting strategies such as ensemble refinement and chain-of-thought reasoning. Both are text-only models.

\subsection{Robustness to Missing Modalities}

One of the challenges central to MEDIQA Task~1~\cite{mediqa-core-task-1} is that for any given case, one or more modalities may be missing. As such, any reasonable approach must handle missing modalities in a robust manner.

One approach is to train a single representation that is tolerant to dropped inputs. HeMIS~\cite{Havaei2016HeMIS} does this with modality-specific CNN encoders, fusing them via the mean and variance over modality feature maps in a shared latent space, yielding a simple, robust hetero-modal segmenter that accepts any subset of modalities without retraining. Similarly, U-HVED~\cite{Dorent2019HVED} embeds all observed modalities into a shared latent that drives both completion and segmentation.

Other approaches push the network to recover missing-modality information. ACN~\cite{Wang2021ACN} trains separate models per missing scenario and uses adversarial co-training between full- and missing-modality models, whereas RFNet~\cite{Ding2021RFNet} uses a per-modality segmentation regularizer to enforce discriminative encoders and mitigate training imbalance.

A third approach uses random masking during training as a regularizer, i.e., simple modality dropout. Hussen-Abdelaziz et al.~\cite{Abdelaziz2020ModalityDropout} applied modality dropout as a regularization scheme for multimodal networks, demonstrating large subjective gains in audio-visual talking-face realism relative to video-only and audio-visual models without dropout. Gu et al.~\cite{Gu2025ContrastiveMultimodalFusion} extended this idea by combining modality dropout with contrastive multimodal learning using learnable modality tokens, achieving state-of-the-art performance on pulmonary embolism detection and lung-cancer prediction with frozen CT encoders and tabular models.

\section{Methodology}

\subsection{Data}

We use the CoRe-BT dataset~\cite{CoRe-BT-arXiv}, which integrates MRI, 
whole-slide histopathology, and diagnostic reports for fine-grained glioma 
subtype classification. It was released as part of Task~1 (Brain Tumor Subtype 
Classification) of the ImageCLEFmed MEDIQA-CORE 2026 challenge~\cite{mediqa-core-task-1,ImageCLEF2026}. The full cohort 
comprises 310 glioma patients recruited at the University of Washington over a 
two-year period (August 2023 to August 2025).

From this cohort, the organizers release three splits: a training set 
($n=183$), a validation set ($n=42$), and a test set ($n=36$). For the additional robustness evaluation, the organizers later introduced a second test set, Test~\#2 ($n=18$), an MRI-only cohort used to evaluate generalization to subjects for whom histopathology was never collected. We refer to the original test set as Test~\#1 ($n=36$) accordingly. Ground-truth labels are provided only for the training set. Validation and test labels are withheld for evaluation. Not every patient has all three modalities. Of the 183 training subjects, 161 (91\%) have MRI, 51 (29\%) have histopathology, and 173 (98\%) have radiology reports. Table~\ref{tab:data-distribution} reports the full MRI/histopathology availability breakdown across all splits.

\begin{table}[!htbp]
    \centering
    \caption{Distribution of MRI/histopathology availability across data splits. ``Both Present'' denotes subjects with both MRI and histopathology; ``Histopathology Only'' and ``MRI Only'' denote subjects missing the other modality.}
    \label{tab:data-distribution}
    \begin{tabular}{lcccc}
        \toprule
        Split & Both Present & Histopathology Only & MRI Only & Total \\
        \midrule
        Train & 35 & 16 & 132 & 183 \\
        Validation & 8 & 3 & 31 & 42 \\
        Test~\#1 & 36 & 0 & 0 & 36 \\
        Test~\#2 & 0 & 0 & 18 & 18 \\
        \midrule
        Full Dataset & 79 & 19 & 181 & 279 \\
        \bottomrule
    \end{tabular}
\end{table}
 
Table~\ref{tab:data-distribution} shows that the two official test sets probe histopathology availability in two different ways. Test~\#1 consists entirely of subjects who have both MRI and histopathology, which is what allows the organizers to artificially withhold either modality at inference time in the ablations of Table~\ref{tab:official-ts1}. In contrast, Test~\#2 consists entirely of MRI-only subjects. Histopathology was never collected for any subject in this split. Comparing performance across the two test sets lets us distinguish a model's robustness to a modality being artificially withheld from its ability to generalize to a population for whom that modality was never available in the first place, and helps identify which modality drives our system's predictions.

\subsection{Baseline System}

The organizers provide a baseline against which submissions are compared. The baseline operates on two of the three modalities, MRI and histopathology embeddings, and does not use the 
radiology reports. 

It is built in two stages. First, a separate linear probe is trained for each 
modality: a single linear layer mapping the 768-dimensional MRI or 
histopathology embedding directly to the per-task class logits. These 
modality-specific probes are then frozen and combined through a gated fusion 
module.

The fusion module combines the two frozen probes in three steps. First, a 
small gating network reads both modality features and produces two weights 
that sum to one, indicating how much to trust each modality. The final 
prediction is the weighted sum of the two probes' outputs. If a modality is 
missing for a subject, its weight is forced to zero so that only the available 
modality contributes. Second, a small residual network produces a learned 
correction term from the combined features. Third, this correction is added to the weighted prediction through a learned scaling factor that is initialized 
near zero. 

As a result, the model begins as a simple weighted combination of 
the two frozen probes and learns only small corrections on top, which 
stabilizes training. The baseline is trained with stochastic gradient descent.

\subsection{Our Approach}

Our approach departs from this baseline in two main ways. First, we incorporate 
the radiology-report modality, giving the model access to all three modalities. Second, rather than learning a single modality gate shared across all three tasks, we explore fusion strategies that let each task draw on the modalities differently.

Concretely, we model MEDIQA-CORE Task~1 as a joint, multi-task classification 
problem over three pre-extracted modalities: (i) MRI volume embeddings, (ii) whole-slide histopathology embeddings, and (iii) radiology-report text embeddings. We explore two fusion architectures: a cross-modal self-attentive fusion and a task-specific gating model, the latter being our primary submitted 
system.

\subsubsection{Cross-Modal Self-Attentive Fusion}

\begin{figure}[!htbp]
    \centering
    \includegraphics[width=0.75\linewidth]{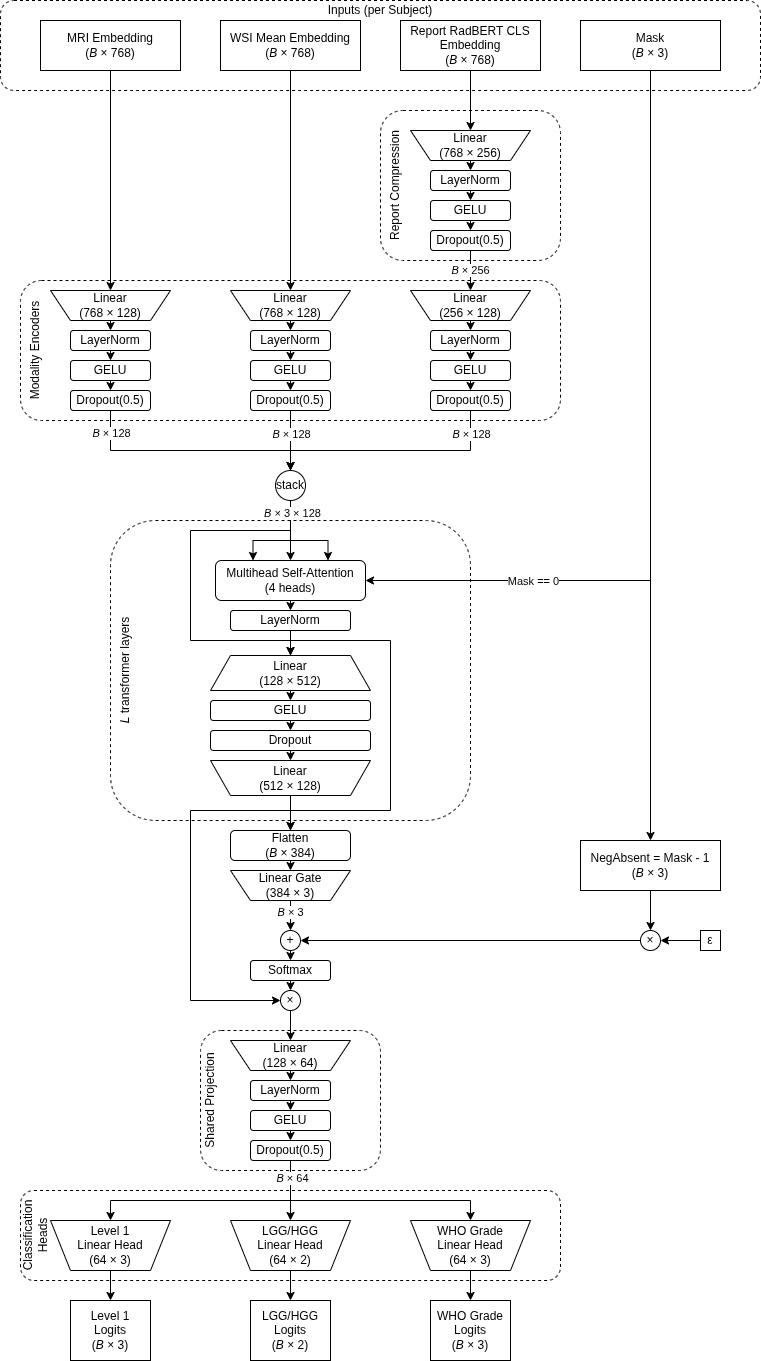}
    \caption{Our cross-modal self-attentive fusion model's architecture.}
    \label{cm-sa-fusion}
\end{figure}

In this model architecture, a shared trunk fuses the three modalities through cross-modal self-attention and a learned modality gate; three independent linear heads predict the targets in parallel. We train with stratified $K$-fold cross-validation and ensemble the $K$ fold models at inference. The full pipeline is summarized in Figure~\ref{cm-sa-fusion}.

\paragraph{Inputs and Pre-Processing}

\subparagraph{MRI and Histopathology.} We use the MRI and pathology embedding ZIPs released by the task organizers. Each MRI embedding is a 768-dimensional vector per subject; histopathology is supplied as one or more 768-dimensional whole-slide-image (WSI) vectors per subject. For subjects with multiple WSI vectors, we aggregate by element-wise mean to a single 768-dimensional histopathology representation.

\subparagraph{Report Embeddings.} We pre-extract a single fixed-dimensional report vector per subject from the redacted radiology reports using the StanfordAIMI/RadBERT checkpoint~\cite{yan2022radbert}, a RoBERTa-base~\cite{liu2019robertarobustlyoptimizedbert} model further pre-trained on radiology corpora. RadBERT is run in full precision; each report is tokenised with the BERT~\cite{devlin2019bertpretrainingdeepbidirectional} tokenizer (max sequence length 512 tokens, including [CLS] and [SEP]). The [CLS] token hidden state from the final encoder layer is used as the report representation (768-dimensional). For reports exceeding 512 tokens, we split the body into overlapping windows of size $512-2=510$ tokens with a stride of 128 fresh tokens between window starts, re-prepend [CLS]/[SEP] per window, run each window through the encoder independently, and mean-pool the per-window pooled embeddings. Pre-extracted embeddings are cached to disk in the same per-subject zip format as the organizer-provided modalities, making them a drop-in third modality at training time.

\subparagraph{Subject Filtering.} We exclude all subjects with level1\_label == 3 (the "Other / NEC" molecular subtype, $n=6$) from training, validation, and the held-out evaluation pool. This class is present in the training metadata but is not enumerated in the public task description, and including it consistently hurt macro-F1 on the Level-1 task in preliminary experiments. After this filter, 177 of 183 training subjects remain.

\subparagraph{Class Weighting.} We compute class weights $w_c = \max_{c'} \frac{n_{c'}}{{n_c}}$ over the training distribution after the filters above. The values we used are provided in Table \ref{cm-sa-fusion-clsweights}.

\begin{table}[!htbp]
    \centering
    \caption{The class weights we used for our cross-modal self-attentive model, broken down by task.}
    \label{cm-sa-fusion-clsweights}
    \begin{tabular}{|c|c|c|}
        \hline
        Task & Number of Classes & Class Weights \\
        \hline\hline
        Level 1 & 3 & ${1.00, 1.60, 7.77}$ \\
        LGG vs HGG & 2 & ${4.23, 1.00}$ \\
        WHO Grade & 3 & ${3.43, 4.29, 1.00}$ \\
        \hline
    \end{tabular}
\end{table}

\subparagraph{Per-Modality Encoders.} Each modality $m \in \{\text{MRI}, \text{Histopathology}, \text{Report}\}$ has its own encoder $f_m: \mathbb{R}^{D_m} \to \mathbb{R}^{D}$ realized as $Linear \rightarrow LayerNorm \rightarrow GELU \rightarrow Dropout$, mapping the input embedding dimension $D_m$ to a shared latent dimension $D = 128$. For the report modality, we first apply a learned bottleneck $Linear(D_R, 256) \rightarrow LayerNorm \rightarrow GELU \rightarrow Dropout$ to equalize the report encoder's capacity with the MRI and histopathology encoders, which already start at $D_R$ comparable to $D_M, D_H$ in our setup but for which we retain the bottleneck for architectural consistency across encoder backbones we ablated.

\subparagraph{Cross-Modal Self-Attention.} Let $z_M, z_H, z_R \in \mathbb{R}^{D}$ denote the resulting per-sample latents. We stack the three latents as a length-3 token sequence $Z = [z_M, z_H, z_R] \in \mathbb{R}^{3 \times D}$ and pass it through $L = 2$ pre-norm transformer encoder blocks. Each block applies multi-head self-attention ($h = 4$ heads) followed by a position-wise feed-forward network with hidden dimension $4D$.

Absent modalities are masked from attention via PyTorch's key\_padding\_mask so they contribute neither queries nor keys/values for their token position.

\subparagraph{Modality Gate.} A learned linear gate $g: \mathbb{R}^{3D} \to \mathbb{R}^{3}$ takes the flattened attended sequence as input and produces three logits, one per modality. Absent modalities have their corresponding gate logit set to $-10^{9}$ before the softmax so that their post-softmax weight is exactly zero. Writing $\alpha = \text{softmax}(g(\text{flatten}(Z)))$, the fused representation is the gated weighted sum of the attended tokens.

\subparagraph{Heads.} A shared projection $Linear \rightarrow LayerNorm \rightarrow GELU \rightarrow  Dropout$ maps $z \in \mathbb{R}^{D}$ to $h \in \mathbb{R}^{P}$ with $P = 64$. Three parallel linear heads then produce the per-task logits: $\text{head}_{L1}: \mathbb{R}^{P} \to \mathbb{R}^{3}$ for Level 1 (after dropping NEC), $\text{head}_{LGG}: \mathbb{R}^{P} \to \mathbb{R}^{2}$ for LGG/HGG, and $\text{head}_{WHO}: \mathbb{R}^{P} \to \mathbb{R}^{3}$ for WHO grade. Dropout with rate $p = 0.5$ is applied throughout. The total trainable parameter count is approximately $8.5 \times 10^{5}$.

\paragraph{Training Regime}

We jointly minimize the unweighted sum of three class-weighted cross-entropy losses with label smoothing $\epsilon = 0.1$:
$$
\mathcal{L} = \mathcal{L}_{L1} + \mathcal{L}_{LGG} + \mathcal{L}_{WHO}.
$$

\subparagraph{Modality Dropout.} As a regularization strategy and to make the model robust to the partial observability typical of the dataset (only $\sim 28\%$ of training subjects have histopathology embeddings), we apply per-sample modality dropout during training. Each modality present for the subject is independently dropped with probability $p_{\text{drop}} = 0.5$; if all three would be dropped, we resample. Dropped modalities are replaced by zero tensors and their mask entry is set to 0, propagating through both the self-attention key\_padding\_mask and the gate's softmax mask.

\subparagraph{Optimizer and Schedule.} We use AdamW with learning rate $1 \times 10^{-4}$, weight decay $1 \times 10^{-3}$, and cosine annealing to $\eta_{\min} = 1 \times 10^{-6}$ over 5{,}000 iterations. Gradients are clipped to $\ell_2$-norm 5.0. The training loader is wrapped in itertools.cycle so iteration count rather than epoch count controls the schedule.

\subparagraph{Validation and Early Stopping.} Every 500 iterations we evaluate on the held-out fold's validation set, computing macro-F1 per task and the mean of the three. The mean macro-F1 is the model-selection metric; the best checkpoint per fold is retained. Training stops early if no improvement is observed over 5 successive validation evaluations (patience = 5).

\paragraph{Cross-Validation and Ensembling}

We first hold out a stratified 10\% subset of the training subjects (stratified on Level-1 label) as an internal test set for sanity-checking the ensembled predictions. On the remaining 90\%, we run stratified 5-fold cross-validation (stratification on Level-1) with seed 42, training one model per fold according to Section~3.4. This yields five fold models with disjoint validation subsets.

At inference time, the five fold models form a logit-averaged ensemble: for each evaluation subject, we average the per-task logits across the five models and then take the argmax to produce the final predictions. Modality dropout is disabled at inference ($p_{\text{drop}} = 0$).

\paragraph{Hierarchical Post-Processing}

The three target tasks form a known label hierarchy: a confident Level-1 prediction implies the LGG/HGG class, and a confident LGG/HGG prediction constrains the WHO-grade prediction (LGG corresponds to grade 0; HGG corresponds to grades $\in {1, 2}$). We apply two soft confidence-thresholded rules on the ensemble's averaged softmax probabilities (threshold $\tau = 0.80$):

If $\hat{p}_{L1} = 0$ (the Astrocytoma class) and $\max \text{softmax}(\hat{p}_{L1}) \geq \tau$, override $\hat{p}_{LGG} \leftarrow 1$.
If $\hat{p}_{LGG} = 0$ (LGG) and $\max \text{softmax}(\hat{p}_{LGG}) \geq \tau$, override $\hat{p}_{WHO} \leftarrow 0$. If $\hat{p}_{LGG} = 1$ (HGG) and confidence $\geq \tau$, mask the WHO=0 logit to $-\infty$ before argmax, forcing the prediction into ${1, 2}$.

These rules act only on high-confidence Level-1 / LGG-HGG predictions, leaving low-confidence cases to the unmodified head outputs.

\paragraph{Implementation and Hyperparameter Summary}

The system is implemented in PyTorch 2.x. All experiments use a single NVIDIA V100 GPU. Random seeds for Python, NumPy, and PyTorch are fixed to 42; cuDNN is set to deterministic mode. Table~\ref{cm-sa-fusion-hyperparams} summarizes the hyperparameters used to produce the submitted prediction.

\begin{table}[!htbp]
    \centering
    \caption{Hyperparameter values used for our Cross-Modal Self-Attentive Model.}
    \label{cm-sa-fusion-hyperparams}
    \begin{tabular}{|c|c|}
        \hline
        Hyperparameter & Value \\
        \hline\hline
        Optimizer & AdamW \\
        Learning Rate & $1\times10^{-4}$ \\
        Weight Decay & $1\times 10^{-3}$ \\
        Learning Rate Schedule & Cosine Annealing to $10^{-6}$ \\
        Iterations (max) & $5,000$ \\
        Batch Size & 32 \\
        Latent Dimension $D$ & 128 \\
        Projection Dimension $P$ & 64 \\
        Dropout & $0.5$ \\
        Modality Dropout $p_{\text{drop}}$ & $0.5$ \\
        Report Bottleneck & $256$ \\
        Number of Self-attention Blocks $L$ & 2 \\
        Number of Self-attention Heads $H$ & 4 \\
        Gradient Clip & $5.0$ \\
        Label Smoothing $\epsilon$ & $0.1$ \\
        Number of Cross-Validation Folds $K$ & 5 \\
        Internal Test Split & 10\% (stratified on Level 1) \\
        Post-processing Threshold $\tau$ & $0.80$ \\
        \hline
    \end{tabular}
\end{table}

\subsection{Trimodal Fusion with Task-Specific Gates}

\begin{figure}[!htbp]
    \centering
    \includegraphics[width=0.75\linewidth]{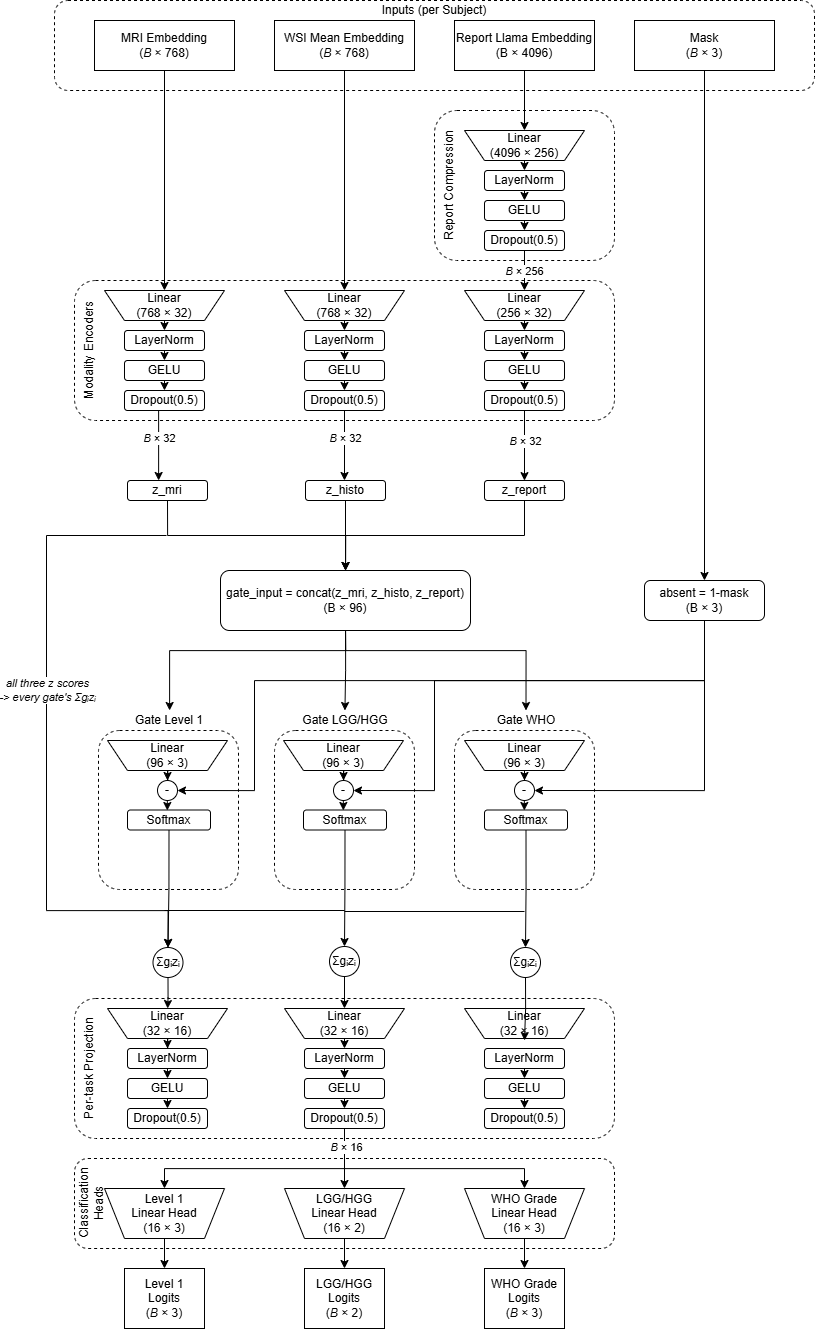}
    \caption{Our trimodal fusion with task-specific gates model's architecture.}
    \label{task_gates}
\end{figure}

While the cross-modal model fuses all three modalities into a single shared 
representation, our second approach asks whether each task might be better 
served by its own fusion. 

The Trimodal Fusion with Task-Specific Gates model is our primary submitted system. It reuses the input preprocessing, subject filtering, class weighting, training regime, cross-validation ensembling, and post-processing described for the cross-modal model. The two systems differ in only two ways: the report encoder and the fusion mechanism. The cross-modal model lets the three modalities interact through self-attention and then combines them with a single shared gate. In contrast, this model gives each task its own gate and does not mix the modalities through attention at all.

\paragraph{Report Embeddings}
Rather than RadBERT, we extract report embeddings with 
Llama-3.1-8B-Instruct~\cite{grattafiori2024llama} as a frozen feature 
extractor. Each report is passed through the model in a single forward pass, 
and the final-layer hidden states are mean-pooled over non-padding tokens to 
produce a 4096-dimensional embedding per report.

\paragraph{Per-Modality Encoders}
Each subject contributes up to three embeddings: a 768-dimensional MRI vector, 
a 768-dimensional histopathology vector, and a 4096-dimensional report vector. 
The report embedding is first compressed to 256 dimensions through a bottleneck 
linear layer. All three vectors are then projected into a shared 
32-dimensional latent space. Each is passed through its own encoder block (Linear 
$\rightarrow$ LayerNorm $\rightarrow$ GELU $\rightarrow$ Dropout). We apply 
dropout $p = 0.5$ throughout.

\paragraph{Task-Specific Gates}
Rather than a single gate shared across tasks, we learn a separate gate for 
each of the three tasks. Each gate is a linear layer that takes the 
concatenation of the three modality latents and produces three softmax weights. These weights
indicate how much each modality contributes to that task's fused 
representation. If a modality is missing for a subject, we set its gate weight to zero so it does not contribute to the fusion. This yields three task-specific fused representations, one per task.

\paragraph{Heads}
Each task's fused representation first passes through its own small processing 
block, a linear layer with LayerNorm, GELU activation, and dropout, which 
reduces it to 16 dimensions. From there, a final linear layer produces the 
class scores for that task: 3 for Level~1, 2 for LGG/HGG, and 3 for WHO Grade. 
Altogether, the model has roughly 280K trainable parameters.

\section{Results}

The organizers evaluate submissions on two test sets that probe different aspects of modality availability. Test Set~1 ($n=36$) has every subject with all three modalities, and the organizers evaluate our model under artificial modality-dropping conditions. Test Set~2 ($n=18$) is a cohort of subjects for whom Pathology was never collected. Together, these two test sets let us distinguish a model's ability to handle a modality that is deliberately withheld at inference from its ability to generalize to a population for whom that modality was never available in the first place.

We report results in two stages: pre-code-verification leaderboard scores, and the organizers' official post-verification scores. The first stage, shown in Table~\ref{tab:results}, reports the pre-verification leaderboard scores on Test Set~1 under the Fully Multimodal condition. The second stage, shown in Tables~\ref{tab:official-ts1} and~\ref{tab:official-ts2}, reports the post-verification scores for Test Set~1 and Test Set~2 respectively, which the organizers obtained by independently re-running our submitted code across all modality conditions.

\subsection{Pre-Verification Results}

We evaluate both of our fusion models on the test set 
and compare them against the organizers' baseline. Each task is scored with 
macro-F1, and the overall ranking metric is the mean of the three per-task 
scores. Table~\ref{tab:results} reports the results.

\begin{table}[!htbp]
    \centering
    \caption{Pre-verification leaderboard Macro-F1 scores under the Fully Multimodal condition for our two systems and the baseline on Test Set~1. Best score per column in bold.}
    \label{tab:results}
    \begin{tabular}{lcccc}
        \toprule
        System & Level-1 & WHO Grade & LGG vs HGG & Mean \\
        \midrule
        Baseline~\cite{mediqa-core-task-1}        & 0.539 & \textbf{0.726} & \textbf{0.920} & 0.728 \\
        Cross-Modal Self-Attentive   & 0.792 & 0.679 & 0.765 & 0.745 \\
        Task-Specific Gates (best score) & \textbf{0.915} & 0.697 & 0.859 & \textbf{0.824} \\
        \bottomrule
    \end{tabular}
\end{table}

Both of our systems outperform the baseline on mean macro-F1. The 
task-specific gating model reaches 0.824, and the cross-modal self-attentive 
model reaches 0.745. Both exceed the baseline's 0.728. The task-specific 
gating model is our strongest system overall. It was our primary submission, 
ranking 4th of 5 teams on the final leaderboard.

The per-task breakdown is more nuanced. Both of our systems improve 
dramatically over the baseline on Level-1 Molecular Type, the hardest and most 
fine-grained task. The task-specific gating model reaches 0.915 and the 
cross-modal model reaches 0.792, compared to the baseline's 0.539. The baseline does not use reports, and neither of our systems was designed to isolate which added modality drove this gain. We further explore this question in Section~4.2 to understand the role of each modality in our predictions.

The baseline, however, remains strongest on the two grade-based tasks. It 
achieves 0.920 on LGG vs HGG and 0.726 on WHO Grade, and neither of our systems 
matches it there. This pattern suggests that reports and our fusion strategies 
help molecular subtyping more than tumor grading. It also suggests that the 
baseline's frozen linear probes are already well-suited to the grade-based 
tasks.

Comparing our two systems directly, the task-specific gating model outperforms 
the cross-modal self-attentive model on all three tasks (Level-1: 0.915 vs 
0.792; LGG vs HGG: 0.859 vs 0.765; WHO Grade: 0.697 vs 0.679). This supports 
our central design choice. Maintaining a separate gate per task, rather than a 
single shared representation, lets each task learn its own modality weighting. 
This yields consistently better predictions. The gap is largest on Level-1 
Molecular Type, the task where the three modalities are most likely to 
contribute unequally.

\subsection{Post-Verification Results}

On Test Set~1, our system ranks second under the Fully Multimodal condition (Table~\ref{tab:official-ts1}). Although our macro-F1 mean of 0.801 is higher than the baseline's 0.796, this lead comes entirely from Level-1, where we score 0.867 against the baseline's 0.672. In turn, the baseline scores better on WHO Grade at 0.795 against our 0.737 and on LGG vs HGG at 0.920 against our 0.800. This split suggests our task-specific gates learn a different modality weighting for Level-1 than for the two grading tasks rather than one fusion strategy serving all three equally well, consistent with the task-specific gating design. 

Dropping Pathology reverses this pattern entirely. The baseline now scores better than our system on all three task metrics and on the mean macro-F1. Our system's mean falls to 0.642 against the baseline's 0.744. This indicates our system relies heavily on Pathology, and that removing it costs us more than it costs the baseline.

When Reports are dropped instead, our system outperforms the baseline on three of the four metrics. It scores 0.841 on Level-1 against the baseline's 0.534, 0.800 on LGG vs HGG against the baseline's 0.438, and 0.720 on mean macro-F1 against the baseline's 0.512. The baseline still scores higher on WHO Grade at 0.564 against our 0.519. Our system withstands the removal of Reports far better than the removal of Pathology, which suggests Pathology carries more of the decision-relevant signal for our gates, even though Reports was the modality we added over the baseline.

Dropping MRI leaves the baseline ahead on three of four columns, with our system scoring higher only on Level-1 at 0.873 against the baseline's 0.672. Taken together, these three ablations show our system's advantage over the baseline is conditional on Pathology being available, rather than a general improvement from adding a third modality. This is a meaningfully different story than a simple three-modalities-beat-two-modalities narrative.

On Test Set~2 (Table~\ref{tab:official-ts2}), the baseline scores higher than our system on every metric in both the MRI+Reports and Reports Only conditions. Our system's mean macro-F1 reaches only 0.493 against the baseline's 0.797 in the MRI+Reports condition, and only 0.229 against the baseline's 0.744 in the Reports Only condition. This cohort has no Pathology available under any condition, so this result is consistent with our earlier finding that our system depends heavily on Pathology. Without it, our system falls well behind a baseline that was designed without Pathology at all.

In the MRI Images Only condition, however, our system scores higher on three of the four metrics. The baseline's advantage on Level-1 is small, at 0.373 against our 0.368. Our system scores the same mean macro-F1 of 0.493 in both the MRI+Reports and MRI Images Only conditions. This finding indicates our predictions for this cohort do not depend on whether Reports are available. The baseline's mean macro-F1, in contrast, falls from 0.797 to 0.405 when Reports are removed, indicating the baseline relies on Reports for this cohort in a way our system does not. This suggests our task-specific gates may have learned to down weight Reports for subjects without Pathology, rather than learning to use Reports as a genuine substitute when Pathology is unavailable.

\begin{table}[!htbp]
    \centering
    \caption{Post-verification Macro-F1 on Test Set~1, across all four modality-availability conditions, for our submitted system (Task-Specific Gates) and the baseline. Best score per column within each condition in bold. Rank is out of all verified teams.}
    \label{tab:official-ts1}
    \resizebox{\linewidth}{!}{%
    \begin{tabular}{llcccc}
        \toprule
        Condition & System & Level-1 & WHO Grade & LGG vs HGG & Mean (Rank) \\
        \midrule
        \multirow{2}{*}{Fully Multimodal}
            & DSGT (Task-Specific Gates) & \textbf{0.867} & 0.737 & 0.800 & \textbf{0.801 (2)} \\
            & Baseline~\cite{mediqa-core-task-1} & 0.672 & \textbf{0.795} & \textbf{0.920} & 0.796 (3) \\
        \midrule
        \multirow{2}{*}{Drop Pathology (MRI + Reports)}
            & DSGT (Task-Specific Gates) & 0.455 & 0.610 & 0.859 & 0.642 (5) \\
            & Baseline & \textbf{0.587} & \textbf{0.726} & \textbf{0.920} & \textbf{0.744 (2)} \\
        \midrule
        \multirow{2}{*}{Drop Reports (MRI + Pathology)}
            & DSGT (Task-Specific Gates) & \textbf{0.841} & 0.519 & \textbf{0.800} & \textbf{0.720 (4)} \\
            & Baseline & 0.534 & \textbf{0.564} & 0.438 & 0.512 (5) \\
        \midrule
        \multirow{2}{*}{Drop MRI (Reports + Pathology)}
            & DSGT (Task-Specific Gates) & \textbf{0.873} & 0.676 & 0.765 & 0.772 (3) \\
            & Baseline & 0.672 & \textbf{0.795} & \textbf{0.920} & \textbf{0.796 (2)} \\
        \bottomrule
    \end{tabular}%
    }
\end{table}

\begin{table}[!htbp]
    \centering
    \caption{Post-verification Macro-F1 on Test Set~2, across all three modality-availability conditions, for our submitted system (Task-Specific Gates) and the baseline. Best score per column within each condition in bold. Rank is out of all verified teams.}
    \label{tab:official-ts2}
    \resizebox{\linewidth}{!}{%
    \begin{tabular}{llcccc}
        \toprule
        Condition & System & Level-1 & WHO Grade & LGG vs HGG & Mean (Rank) \\
        \midrule
        \multirow{2}{*}{MRI + Reports}
            & DSGT (Task-Specific Gates) & 0.368 & 0.439 & 0.673 & 0.493 (5) \\
            & Baseline~\cite{mediqa-core-task-1} & \textbf{0.816} & \textbf{0.716} & \textbf{0.858} & \textbf{0.797 (1)} \\
        \midrule
        \multirow{2}{*}{MRI Images Only}
            & DSGT (Task-Specific Gates) & 0.368 & \textbf{0.439} & \textbf{0.673} & \textbf{0.493 (4)} \\
            & Baseline & \textbf{0.373} & 0.409 & 0.433 & 0.405 (5) \\
        \midrule
        \multirow{2}{*}{Reports Only}
            & DSGT (Task-Specific Gates) & 0.127 & 0.127 & 0.433 & 0.229 (4) \\
            & Baseline & \textbf{0.808} & \textbf{0.567} & \textbf{0.858} & \textbf{0.744 (1)} \\
        \bottomrule
    \end{tabular}%
    }
\end{table}

\section{Future Work}

Our results point to several directions for future work. While our task-specific gating model performed well on molecular subtyping, the gap on the grade-based tasks and our limited use of the available data leave clear room for improvement.

\paragraph{Closing the Grade-Task Gap}
The baseline outperformed both of our systems on LGG vs HGG and WHO Grade, 
despite using fewer modalities. This suggests that adding the report modality 
and learning per-task gates helped molecular subtyping but not tumor grading. 
A natural next step is to investigate why. One possibility is that the 
grade-based tasks rely more heavily on imaging features that our fusion 
dilutes. Another is that the frozen probes in the baseline are simply better 
calibrated for these tasks. Increasing the loss weight on the WHO Grade and 
LGG/HGG heads, or training a dedicated model for the grade-based tasks, may 
help recover the lost performance.

\paragraph{Tile-Level Histopathology Features}
The organizers release histopathology as slide-level mean-pooled embeddings. 
The test set additionally includes tile-level features that we did not use, 
since they were incompatible with our slide-level training format. Future work 
could explore learned aggregation from tile embeddings, such as attention-based 
pooling, which may preserve fine-grained cellular detail relevant to tumor 
grading.

\paragraph{Improving Generalization to Histopathology-Absent Subjects}
The official evaluation on Test Set~2 shows our system trails the baseline substantially in a cohort where histopathology was never collected, even though our system depends heavily on histopathology when it is present. This suggests our task-specific gates may have learned to rely on histopathology-correlated patterns that do not transfer to a genuinely histopathology-absent population, rather than learning a robust MRI/Reports-only decision path. Because only around 29\% of our training subjects have histopathology, increasing the modality dropout rate specifically for histopathology during training, or training a dedicated MRI/Reports-only variant of the gating network, are next steps to close this gap.

\section{Conclusions}

We presented the DS@GT ARC team's submission to Task~1 of the ImageCLEFmed 
MEDIQA-CORE 2026 challenge ~\cite{mediqa-core-task-1,ImageCLEF2026}. The task jointly evaluates three glioma classification 
tasks: Level-1 Molecular Type, LGG vs HGG, and WHO Grade. Building on the 
pre-extracted MRI and histopathology embeddings released by the organizers, we 
added a third modality, free-text radiology reports. We explored two trimodal fusion architectures: a cross-modal self-attentive model and a task-specific gating model.

The two architectures differ in how they combine the modalities. Our 
stronger model learns a separate gate for each task, rather than fusing 
everything into one shared representation, so each task can weight the 
modalities independently. This design consistently scored higher than the cross-modal model on all three tasks. The finding supports the idea that the tasks benefit from different modality combinations.

Our best submission achieved a mean macro-F1 of 0.801 under the Fully Multimodal condition, ranking 2nd among the teams whose code passed verification and exceeding the organizers' baseline of 0.796. This advantage was driven primarily by a large gain on Level-1 Molecular Type, where we reached 0.867 compared to the baseline's 0.672. On the grade-based tasks, however, the baseline remained stronger. Our official per-modality ablation further shows that this advantage depends heavily on the histopathology modality rather than on the report modality we added over the baseline, and does not hold when histopathology is unavailable, whether withheld at inference or absent from the cohort entirely.

\begin{acknowledgments}
We thank the Data Science at Georgia Tech (DS@GT) CLEF competition group for their support. This research was supported in part through research cyberinfrastructure resources and services provided by the Partnership for an Advanced Computing Environment (PACE) at the Georgia Institute of Technology, Atlanta, Georgia, USA~\cite{PACE}.
\end{acknowledgments}

\section*{Declaration on Generative AI}
 
  During the preparation of this work, the author(s) used Claude in order to: Grammar and spelling check, Content brainstorming, and writing data-analysis and evaluation scripts. After using these tool(s)/service(s), the author(s) reviewed and edited the content as needed and take(s) full responsibility for the publication's content.

\bibliography{main}
\end{document}